\documentclass[10pt,journal,cspaper]{IEEEtran}
\ifCLASSOPTIONcompsoc
  \usepackage[nocompress]{cite}
\else
  \usepackage{cite}
\fi

\usepackage{booktabs}
\usepackage{cite}
\usepackage{amsmath,amssymb,amsfonts}
\usepackage{algorithmic}
\usepackage{textcomp}
\usepackage[ruled]{algorithm2e}
\usepackage{multirow}
\usepackage{float}
\usepackage{makecell}
\usepackage{stfloats}
\usepackage{hyperref}
\usepackage{xcolor}
\usepackage{url}

\newtheorem{lemma}{Lemma}

\newtheorem{proposition}{Proposition}[section]

\usepackage{ragged2e}

\usepackage{subcaption}
\usepackage[pdftex]{graphicx}

\ifCLASSINFOpdf
  \usepackage[pdftex]{graphicx}
\else
  \usepackage[dvips]{graphicx}
\fi

\begin{document}

\title{Is Data Valuation Learnable and Interpretable?}

\author{Ou Wu, Weiyao Zhu, Mengyang Li 
\IEEEcompsocitemizethanks{\IEEEcompsocthanksitem Ou Wu~(Corresponding author), Weiyao Zhu, and Mengyang Li are with the National Center for Applied Mathematics, Tianjin University, Tianjin,
China, 300072. E-mail: \{wuou, wyzhu, limengyang\}@tju.edu.cn\protect\\
}
\thanks{Manuscript received 2024.}}

\markboth{Journal of \LaTeX\ Class Files,~Vol.~XX, No.~X, X~20XX}%
{Shell \MakeLowercase{\textit{et al.}}: Bare Demo of IEEEtran.cls for Computer Society Journals}

\maketitle



\begin{abstract}
\justifying
Measuring the value of individual samples is critical for many data-driven tasks, e.g., the training of a deep learning model. Recent literature witnesses the substantial efforts in developing data valuation methods. The primary data valuation methodology is based on the Shapley value from game theory, and various methods are proposed along this path. {Even though Shapley value-based valuation has solid theoretical basis, it is entirely an experiment-based approach and no valuation model has been constructed so far.} In addition, current data valuation methods ignore the interpretability of the output values, despite an interptable data valuation method is of great helpful for applications such as data pricing. This study aims to answer an important question: is data valuation learnable and interpretable? A learned valuation model have several desirable merits such as fixed number of parameters and knowledge reusability. An intrepretable data valuation model can explain why a sample is valuable or invaluable. To this end, two new data value modeling frameworks are proposed, in which a multi-layer perception~(MLP) and a new regression tree are utilized as specific base models for model training and interpretability, respectively. Extensive experiments are conducted on benchmark datasets. {The experimental results provide a positive answer for the question.} Our study opens up a new technical path for the assessing of data values. Large data valuation models can be built across many different data-driven tasks, which can promote the widespread application of data valuation.
\end{abstract}


\section{Introduction}
\IEEEPARstart{D}{ata} value is among the most important characteristics of big data. Assessing the value of individual data is beneficial or even crucial in many data-driven applications. For example, in data-centric deep learning, data valuation can be used to perceive the training set and then to select/discard useful/useless training samples~\cite{OAU2023}. In data market, the accurate quantifying the data value is critical for the fairness of data transactions~\cite{ASODP2022}, as a fair data pricing mechanism is dependent of accurate data valuation. A large number of data valuation methods have been presented in previous literature. Ghorbani and Zou firstly introduced Shapley value for data valuation~\cite{DSE2019}. It is a concept from the game theory~\cite{AVF1953}, which is one of the most popular measure for data valuation. The calculation for the classical Shapley value is NP-hard, thereby hindering its use in real applications. Yoon et al.~\cite{DVUR2020} proposed a reinforcement learning-based method for data valuation. Nevertheless, their inferred weights reflect the importance of a sample in learning rather than the classical Shapley value for samples.

Recently, Lin et al.~\cite{MIE2022} proposed an approximate yet efficient measure, namely, average marginal effect~(AME), for Shapley value with theoretical guarantee. Although significant progress has been made in previous studies, several unresolved issues remain closely related to Shapley-based data valuation, including:
\begin{itemize}
    \item The number of parameters involved in most existing methods is not fixed. Most Shapley value-based methods take the values of each sample as parameters~(i.e., variables) to infer. Therefore, the number of parameters equals to the number of involved samples. In other words, the number of parameters is not fixed and if the data size is large, then the entire computational complexity will be extremely high.
    \item Knowledge among the training data in the same or similar domains are not reusable. In each existing Shapley value-based methods, the calculation procedures among different tasks are completely unrelated. Consequently, knowledge about data valuation from one task is inapplicable to another task from the same or similar domains. To our knowledge, all existing data valuation studies do not mention this issue. Nevertheless, the knowledge reuse for data valuation will spawn many potential techniques and applications, e.g., large valuation models.
    
    \item Existing data valuation results are not interpretable. As Shapley value is based on the cooperative game experiment, there is no data valuation models generated in all existing methods. Without doubt, the valuation results are not interpretable. In other words, which exactly characters for a sample resulting its high/low value remains totally unknown. Interpretability of the data valuation results is useful or even crucial in many applications. For example, interpretable data valuation is quite important for fair data transactions in data market.
\end{itemize}

This study resorts to machine learning to address the three issues listed above. 
If we can learn a data valuation model, then the three abovementioned issues can be significantly overcome. First, a data valuation model to learn can have a fixed number of parameters even confronting with different tasks. Secondly, a data valuation model learned on a task can be transferred to another task from the same/similar domain. Hence, knowledge is reused between the two tasks. Third, at least two strategies can be applied to obtain interpretable valuation results: 1) interpretable technology can be applied to the learned model to generate interpretable insights; 2) if the employed base model has interpretability on its own, then the valuation results will naturally be interpretable. With this regard, this study aims to make the first step to explore that whether data valuation is learnable and interpretable.

To answer the question, this study designs a new framework which brings the machine learning paradigm into data valuation and evaluates the framework via extensive experiments on benchmark data corpus. Specifically, a series of features which can characterize or reflect a sample's characteristics during model training are firstly extracted. A new learning framework is then proposed which can seamless integrate existing SOTA data valuation approach and a data valuation model. The data valuation model can thus be trained under the learning framework. This study considers two kinds of data valuation models. First, an MLP network is utilized as the data valuation model. To further enhance the interpretability of the valuation model, a new regression tree, namely, sparse regression tree~(SRT), is proposed which can embedded in the whole learning framework, making the data valuation model interpretable. Extensive experiments on benchmark corpus verify the effectiveness of the proposed two frameworks. The performance of the learning model is similar to those achieved by SOTA methods, indicating that the a data valuation model can be learned. The performance of the proposed SRT is also comparable to SOTA methods and numerious rules can be extracted from the learned SRT, indicating that a interpretable data valuation model can be learned.

Our contributions are summarized as follows:
\begin{itemize}
    \item A learning framework is proposed which can train a concrete data valuation model. This framework can be seamless integrate an existing data valuation method and the training of a data valuation model. To the best of our knowledge, this is the first work that attempts to establish a concrete data valuation model. Compare with existing game experiment-based methods, a data valuation model has valuable merits.
    \item A series of training characteristics which can capture a wide range of aspects of a sample are utilized. The relevance of these features to the data valuation is explored. Multi-layer perceptron~(MLP) is utilized as the base model for valuation. In theoretical, any differentiable regression models can be used.
    \item A sparse regression tree~(SRT) is proposed which can iteratively optimized via a stochastic gradient descent~(SGD)-like learning process. With the trained SRT, the data valuation results are inherently interpretable.
    \item Results from extensive experiments based on our proposed methodology give a positive answer to the question investigated in this study. Accordingly, this study opens up a new direction that aims to construct a model for data valuation.
\end{itemize}

This paper is organized as follows. Section~2 introduces related studies for data valuation. Section~3 describes the proposed methodology including the main learning framework, features, and training procedure. Section~4 introduces the proposed sparse regression tree~(SRT) and SRT-based interpretable data valuation. Section~5 introduces the experimental evaluation and the answers for the question investigated in this paper, and conclusions are presented in Section~6.

\section{Related Work}
\subsection{Data Valuation}
There are various technical paths to achieve the value of a data in machine learning. This study focuses on the Shapley value-based path. The Shapley value for data valuation~\cite{DSE2019} is defined as follows:
\begin{equation} 
\phi(x_i) = \sum_{S \in D-x_i}\frac{1}{C_{|D|-1}^{|S|}} [V(S \cup \{x_i\})-V(S)]\label{SV1}, 
\end{equation}
where $V(\cdot)$ is the utility function of a dataset and $S$ is a subset of the training corpus $D$. Nevertheless, the accurate valuation based on Eq.~(\ref{SV1}) is NP-hard. Therefore, numerous studies explore approximate yet efficient algorithms towards Shapley values. Jia et al.~\cite{TED2019} introduced two practical SV estimation methods specific to ML tasks. The first method relies on the assumption that the employed learning algorithm is uniformly stable; the second method relies on the influence function~\cite{UBP2017} without provable guarantees on the approximation error. Kwon and Zou~\cite{BSA2022} proposed a generalization of Shapley value, namely, Beta Shapley. They also designed an efficient algorithm to estimate Beta Shapely. Most recently, Lin et al.~\cite{MIE2022} proposed a linear regression-based method to approximate the Shapley value. Theoretical guarantee is also established in their study. Jiang et al.~\cite{OAU2023} constructed an easy-to-use and unified framework that facilitates researchers and practitioners to evaluate their proposed data valuation algorithms under different dimensions.

\subsection{Feature Attribution}
Most recent studies leverage feature attribution to  explain a DNN's prediction by assigning scalar values that represent each feature’s importance~\cite{ATE2023}. Shapley value has also been widely used in feature attribution. Mukund and Amir~\cite{TMS2020} leveraged the axiomatic approach to explore the differences between several operationalizations of the Shapley value for feature attribution. Chai et al.~\cite{AFAE2023} utilized shapley values to measure the sensitivity of input features. To explain DNNs, Zhang et al.~\cite{IMS2021} defined and quantified the significance of interactions among multiple input features based on Sharply values. In feature attribution, researchers have developed many efficient calculation methods for Shapley value. Most recently, Chen et al.~\cite{HCA2023} proposed a deep neural network-based method, namely, HarsanyiNet, which can computes the exact Shapley values of input features very quickly.

\subsection{Data Perception}
Data perception aims at capturing training data's intrinsic characteristics and patterns that affect learning performance. Wu and Yao~\cite{wu2023data} summarized eight typical quantifying types used for data perception, including distribution, cleanliness, difficulty, uncertainty, diversity, balance, consistency, and value. These quantities are widely used to identify noisy-label samples or guide the optimize the learning processing for deep learning tasks. For example, one main application of data valuation is noisy-label learning. Sample cleanliness, difficulty, and uncertainty are usually employed to detect noisy labels.

Five types among the eight ones, namely, cleanliness, difficulty, uncertainty, consistency, and value can quantify each sample. The former four types can have explicit calculation formulas and thus they are interpretable. Nevertheless, data value, especially Shapley-based measurement, does not have explicit formulas. Instead, it depends on game experiments and thus it is not interpretable. In other words, we do not know the intrinsic mechanism that determine whether the value of a sample is high or low.

Given that there are interpretable measures which can characterize each sample with different views, it is worth explorting the relationships between data value and existing interpretable measures. This study attempts to model the data value based on existing interpretable measures.

\section{Methodology}\label{Methodology}
Our proposed methodology can be adapted to many existing data valuation methods, which will be discussed in Section 5. In this section, we take the recent SOTA method approximate marginal effect~(AME)~\cite{MIE2022} as an illustrative example to describe our method.

The symbols and notations are defined as follows. Let $\mathcal{D} = \{\boldsymbol{x}_i, y_i\}_{i=1}^n$ represent a set of $n$ training samples, where $\boldsymbol{x}_i$ is the feature and $y_i \in \{1,2,\cdots,C\}$ is the label, and $C$ is the number of categories. Let $n_c$ be the number of the samples in the $c$th category of the training set. Let $\mathcal{S}$ denote the set of all data subsets of $\mathcal{D}$. Let $\beta = \{\beta_1, \beta_2, \cdots, \beta_N\}^T$ be the data values of the $N$ training samples.

\subsection{The AME Valuation Method}\label{AME}
AME~\cite{MIE2022} is inspired by by causal inference and randomized
experiments: different subsets of
the training corpus are sampled to train multiple sub-models; submodel’s performance is evaluated and used to infer the data value of each training sample in the training corpus. It consists of four steps as follows:
\begin{itemize}
    \item[(1)] Training subset sampling. AME firstly samples a probability $p$ from a uniform distribution $p \sim \text{Uni}(0; 1)$. Each training sample is then selected according to $p$ to constitute a new training subset. This process repents until $M$ subsets are obtained.
    \item[(2)] Models training and assessing. For each sampled subset, a model is trained and its performance is evaluated. Let $r_m$ be the performance of the $m$th subset. In a classification problem, the performance measures can be accuracy, $F_1$ measure, or others. Let $R$ be the performance vector of all the $M$ subsets.
    \item[(3)] Feature extraction. In this step, an $M$-dimensional feature vector for each sample is generated. The $m$th feature for $x_i$, denoted as $\hat{x} 
    _{im}$, is defined as follows: $\hat{x} 
    _{im} = \frac{1}{\sqrt{v}p_m}$ if $x_i \in \mathcal{S}_m$ and $\hat{x} 
    _{im} = -\frac{1}{\sqrt{v}(1-p_m)}$ otherwise, where $v = E_p[\frac{1}{p(1-p)}]$ and $p_m$ is the sampling rate for $\mathcal{S}_m$.
    \item[(4)] Value inference. Based on the performance vector $R$ and new features $\hat{X}$, the values of each sample can be inferred by solving the following Least squares problem:
       \begin{equation} 
        \beta^* = arg\underset{\beta}{\min}||R- \hat{X}\beta||_2^2, \label{mse-1}
       \end{equation}
    where $\beta^*_i$ is the inferred value of sample $x_i$.
\end{itemize}
Lin et al.~\cite{MIE2022} proved that if $M \to +\infty$, $\beta^*_i$ is approach to the true Shapley value of $x_i$ with the following two theorems:
\begin{proposition}
    When $M \to +\infty$, $\beta^*_i$ solved from (\ref{mse-1}) is proportion to the AME value of each sample: 
        \begin{equation}
            \beta^*_i =\frac{AME_i}{\sqrt{v}},\label{ame-1}
        \end{equation}
    where $AME_i$ is the average marginal effect of $x_i$.
\end{proposition}
and
\begin{lemma}
    If $p \sim \text{Uni}[0,1]$, then $\text{AME}_i$ equals to the Shapley value of $x_i$. 
\end{lemma}

Although $\beta^*$ seems an ideal approximate for Shapley value, in real applications, the number of $M$ should be not quite large. The reason lies in that when $M$ is large, the computational load for the construction of $M$ models is considerably heavy. Resultantly, if $M$ is sufficiently smaller than the sample size $N$, (\ref{mse-1}) becomes an under-determined regression problem. Lin et al.~\cite{MIE2022} introduced a sparsity assumption that the number of samples with non-zero AME values $k$ is much less than the entire corpus size $N$, namely, $k \ll N$. With this reasonable assumption supported by statistical results on several datasets, the under-determined regression problem can be better addressed with the following LASSO optimization:
        \begin{equation} 
       \Tilde{\beta} = arg\underset{\beta}{\min}||R- \hat{X}\beta||_2^2+\lambda |\beta|, \label{lasso-1}
       \end{equation}
where $\Tilde{\beta}$ is the optimal solution of LASSO, and $\lambda$ is a hyper-parameter. In essential, LASSO compresses the solution space to alleviate the under-determined issue. Theoretical analysis in ~\cite{MIE2022} indicated that when $M \to +\infty$, $\Tilde{\beta} \to \beta^*$. To distinguish the algorithm based on (\ref{mse-1}) with the LASSO above, they are called AME-standard and AME-LASSO, respectively. 

As being analyzed in Section \ref{sec:introduction}, no matter AME-standard and AME-LASSO, they both confront with three limitations. First, AME requires the training of $O(klogN)$~(i.e., $M \sim klogN$) models. If the training of a single model~(such as deep neural network) is time consumption, the entire computation is still prohibitively expensive. Second, even if only one or several new samples are added in the corpus, all the four steps required to be re-execute. The cost is nearly acceptable. Thirdly, the inferred values are not intepretable. That is, we are aware that a large value of $\Tilde{\beta}_i$ means the average marginal effect of sample $x_i$ is high. However, we have completely no sense about why $x_i$ has high average marginal effect.

To address the issues confronting AME and also many existing data valuation methods, the following subsection introduces our attempt on the pursuing of a machine learning-based methodology which aims to train a data valuation model. Moreover, interpretable modeling is also explored in the succeeding part. 
 
\subsection{Overall of The Proposed Method}
Different from all existing Shapley value-based data valuation methods that ignore the characteristics of each involved player in the game, our learning-based methodology assumes that a mapping function or a model exists between a player's characteristics and its Shapley value, namely:
\begin{equation} 
       {\beta}_i = f(u_i;\Theta),\label{LV-1}
\end{equation}
where $u_i$ represents all relevant characteristics for the player $x_i$; $f(\cdot;\Theta)$ is the mapping function or saying a model; and $\Theta$ is the parameter to achieve. In a machine learning application, the characteristics can be all useful information surrounding a sample, such as feature, uncertainty, cleanness, difficulty, the characters of its located class, the characters of the entire corpus, etc.

The assumption of Eq.~(\ref{LV-1}) implies that if two or more samples with very close characteristics, then their Sharpley values should be very similar or even identical. This assumption conforms to an perspective of Causal theory. More over, if a sophisticated yet interpretable base model for $f(\cdot;\Theta)$ is adopted, then the predicted Shapley value is also intepretable.

Let $U = [u_1,u_2,\cdots,u_N]^T$ and $\beta(U;\Theta) = [f(u_1;\Theta),\cdots,f(u_N;\Theta)]^T$. On the basis of (\ref{mse-1}) and Eq.~(\ref{LV-1}), the learning for $f(\cdot)$ is formalized into the following optimization problem:
       \begin{equation} 
        arg\underset{\Theta}{\min}||Y- \hat{X}f(U;\Theta)||_2^2.\label{LV-2}
       \end{equation}

If $f(\cdot;\Theta)$ can really be learned, then compared with the method based on (\ref{mse-1}), our methodology based on (\ref{LV-2}) has the following merits:
\begin{itemize}
    \item The number of variables to be solved can be significantly reduced. The number of variables in (\ref{mse-1}) is $N$. In (\ref{LV-2}), the number of variables is subject to $\Theta$, which can be designed to be far smaller than $N$. Accordingly, the strong sparsity assumption employed for (\ref{lasso-1}) is non longer required. 
    \item Once $f(\cdot;\Theta)$ is obtained, the values of new added samples can be efficiently achieved only requiring the calculation of the characteristics, i.e., $u_i$.
    \item If $f(\cdot;\Theta)$ is a model like decision tree, then which characters resulting in a large/moderate/small value of a player can be deduced.
\end{itemize}

\textcolor{red}{Fig.~1} shows the overall pipeline of our learning-based data valuation framework.

\subsection{Player's Characteristics~($u_{\cdot,i}$)}
The characterizing of a player in terms of its marginal effect is subject to the application scenario. As this study limits the scenario in the classification tasks, we take a classification task as an example to illustrate the extraction of a player's characteristics. Four basic principles are considered as follows:
\begin{itemize}
    \item[(1)] Two normal samples with close features should have close data values.
    \item[(2)] Local and global factors other than the samples themselves should not be ignored such as their neighborhood, located categories, and the entire corpus.
    \item[(3)] A low-dimensional yet effective set of characteristics for each sample is favored.
    \item[(4)] A more universal set of characteristics is favored.
 \end{itemize}
 Regarding the first principle, the emphasis for ``normal" lies in that if two samples close to each other have different labels, their values may differ significantly. One sample among the two may be noisy and thus its effective is harmful~(its value is negative). Regarding the second principle, quantities such as category proportion of a sample also determine its value. For example, each sample in a tail category in long-tailed learning is valuable, whereas many samples in a head category may be redundant and thus useless. Regarding the third and the fourth principles, a lower dimension of input characters reduces the requirement of a complex model with a large number of parameters, and universal input characters is beneficial for the construction of a universal data valuation model across different tasks.

According to the four principles, sample feature is unsuitable for using as input characters. There are two reasons. First, the utilization of feature may broke the second principle due to the existence of noisy-label samples. Second, the feature dimension is high in many applications. We resort to a wide range of studies that capture the characteristics for both standard learning scenario and learning under imperfect training data. Ten quantities~($u_{\cdot,1}, \cdots, u_{\cdot,10}$) are considered in this study:
\begin{itemize}
    \item Loss: It is the most widely used quantity~\cite{SAG2020,SPLA2017} to inspect the training characteristics of a sample. Both the average loss~($u_{\cdot,1}$) and the loss variations~($u_{\cdot,2}$) in a training process are used.
    \item Gradient norm: This quantity is usually used to identify noisy samples. The norm of the gradient on logits is used with the following formula:
       \begin{equation} 
      u_{\cdot,3}= |q-y|,\label{U-1},
       \end{equation}
    where $q$ is the softmax output and $y$ is the one-hot label for a sample. Likewise, the norm variation~($u_{\cdot,4}$) along training epochs is also used.
    \item Uncertainty: The samples near to the class boundary usually have larger uncertainty values. The information entropy-based uncertainty~\cite{PDA2008} is used in this study. Both the average uncertainty~($u_{\cdot,5}$) and its variation~($u_{\cdot,6}$) along training epochs are used. 
    \item Forgetting: This quantity receives much attention in recent deep learning literature~\cite{ACSOF2023}. Forgetting occurs when the sample is predicted correctly in the previous epoch and is wrongly predicted in the current epoch. The forgetting count~($u_{\cdot,7}$) of a sample throughout the training process is recorded.
    \item Neighborhood inconsistency: This quantity~($u_{\cdot,8}$) describes the difference between the label of a sample and the labels of samples in its $k$-nearest neighbors. As the searching of the $k$-nearest neighbors at each epoch is high time consumption, this study adopts a simple method. The model at the epoch corresponding to the best validation performance is recorded. The features output by this model are used to achieve the $k$-nearest neighbors of each sample. The KL-divergence is used. 
    \item Category proportion: This quantity~($u_{\cdot,9}$) is important in imbalanced learning.
    \item Category-wise forgetting: The average forgetting count~($u_{\cdot,10}$) and forgetting variation~($u_{\cdot,11}$) for the category of the sample are calculated and used. These two quantities also reflect the overall characteristics of the category.
\end{itemize}
All the above-mentioned characters are standardized by $z$-score according to two considerations. First, different learning tasks may have different amplitudes on the extracted quantities. The $z$-score standardization can ensure the universality of the involved characters. Second, the $z$-score standardization embeds the information of the entire corpus. In total, 11-dimensional characters are extracted for each sample. They are universal and can be extracted in common classification tasks. 

To demonstrate the usefulness of the ten characteristics, we make a statistic for the relevance between Shapley value and each of the ten characteristics on two benchmark data sets used in our experiments. One set is the standard CIFAR10, and the other is imbalanced dataset CIFAR100-LT. The Shapley value is approximated by AME with a large number of $M$. \textcolor{red}{Fig.~2} shows the feature differences of samples as the feature similarity of the samples changes.

These characters are input into a base model to train a data valuation model. The next subsection introduces MLP-based model.

\subsection{MLP-based Valuation}
Mulyi-layer perception~(MLP) is a classical neural network widely used in many tabular data-oriented learning tasks. According to the universal approximation theory, a three layer can approximate arbitrary continuous functions. It can be iteratively optimized via error back propagation based on stochastic gradient descent~(SGD). Accordingly, (\ref{LV-2}) is specified as follows:
       \begin{equation} 
        arg\underset{\Theta}{\min}||R- \hat{X}f_{MLP}(U;\Theta)||_2^2,\label{LV-MLP}
       \end{equation}
where $f_{MLP}(U;\Theta)$ is the MLP model and $\Theta$ is the parameters. The entire network for (\ref{LV-MLP}) can be illustrated by Fig.~3. $\hat{R}_m$ is the prediction for $\hat{X}_m$, i.e., $\hat{R}_m = \sum_{i=1}^N \hat{x}_{im}\beta_m$. All the MLP modules in the network share all their parameters. 
\begin{figure}[t] 
    \centering \includegraphics[width=1\linewidth]{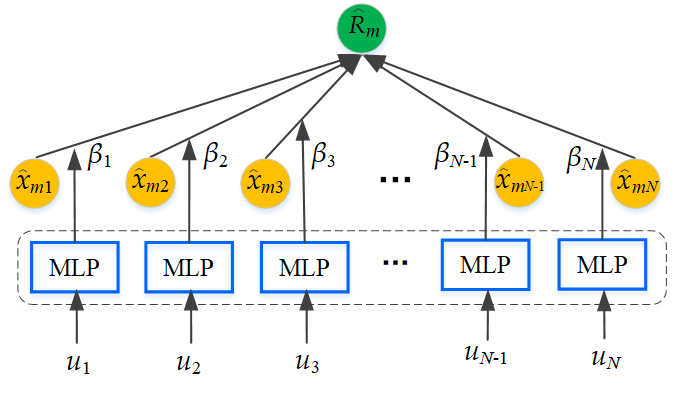}     \vspace{-0.3in}
    \caption{The MLP-based regression for data valuation.}
    \label{fig1}
     \vspace{-0.02in}
\end{figure}

In this study, the size of the hidden layer in the employed tree-layer MLP is chosen from \{50, 75, 100\}. Therefore, the total number of parameters in our MLP-based valuation is about 550, 825, or 1100. The number is fixed no matter the size of the entire data corpus. Therefore, the parameters are significantly reduced compared with those in (\ref{mse-1}) on large data corpora. Subsequently, the strong sparsity assumption used in (\ref{lasso-1}) is no longer required.

The learning for the parameters $\Theta$ still depends on SGD with the following formula:
 \begin{equation} 
       \small
        \begin{aligned}
            \Theta^{t+1} &= \Theta^{t} - \eta\frac{\partial ||R- \hat{X}f_{MLP}(U;\Theta)||_2^2}{\partial \Theta}|_{\Theta^{t}}\\
            &=\Theta^{t} -\eta(\frac{\partial [R- \hat{X}f_{MLP}(U;\Theta)]}{\partial \Theta})^T \\
            &\qquad \times \frac{\partial ||R- \hat{X}f_{MLP}(U;\Theta)||_2^2}{\partial [R-\hat{X}f_{MLP}(U;\Theta)]}|_{\Theta^{t}} \times \frac{\partial \hat{X}f_{MLP}(U;\Theta)}{\partial \Theta}|_{\Theta^{t}}\\
            &=\Theta^{t} +2\eta[\hat{X}\frac{\partial f_{MLP}(U;\Theta)}{\partial \Theta}]^T[R-\hat{X}f_{MLP}(U;\Theta)]|_{\Theta^{t}}        \end{aligned},\label{SGD1}
       \end{equation}
where $\eta$ is the learning rate. Assume that the involved MLP contains $H$ parameters~(i.e., $\Theta \in \mathcal{R}^H$). The term $\frac{\hat{X}f_{MLP}(U;\Theta)}{\partial \Theta}$ is a standard update step for MLP. Its rank is $N \times H$. According to (\ref{SGD1}), for a batch of instances\footnote{Here the instances are not the original samples.} $B$, the update of $\Theta$ under SGD is
      \begin{equation} 
       \small
        \begin{aligned}
            \Theta^{t+1} &= \Theta^{t} + 2\eta \frac{1}{|B|}\sum_{b_M=1}^{|B|}[\hat{x}_{b_M}\frac{\partial f_{MLP}(U,\Theta)}{\partial \Theta}]^T \\
            & \qquad\qquad\qquad\qquad \times [R_{b_M}-\hat{x}_{b_M}^Tf_{MLP}(U,\Theta)]|_{\Theta^{t}}
        \end{aligned},\label{SGD2}
       \end{equation}
where $b_M$ is the index of the $b_M$th instance of $B$ in the entire corpus, and $R_{b_M}$ is the performance of the model trained on the $b_M$th subset. 

In the implementation, the data corpus~($\hat{X}, R$) is randomly divided into five folds and the MLP is trained via five-fold cross validation. The whole algorithm is called MLPbV and the algorithmic steps are presented in Algorithm 1.

\begin{algorithm}[t]
\small
    \caption{MLPbV}
    \label{alg1}
    \textbf{Input}: Original corpus $D= \{X, Y\}$, sampling rate $p$, $M$, K, batch size $|B|$, \#epochs $T$, $\eta$.\\
       \textbf{Output}: Trained valuation network MLP($\cdot$).
    \begin{algorithmic}[1]\vspace{-0.0in}
\STATE {Initialize $\Theta^{0}$;}
\FOR{$m = 1$ to $M$}
    \STATE{Sample a subset $D_m$ from $D$ with probability $p$;}
    \STATE{Train a model based on $D_m$;}
    \STATE{Evaluate the performance of the model and record it as $R_m$;}
\ENDFOR
\STATE {Calculate new feature matrix $\hat{X}$;}
\STATE {Extract the value features $U$ based on an independent learning process on raw samples and labels;}
\STATE {Randomly divide $\{\hat{X}, R, U\}$ into five folds;}
\FOR {$t = 1$ to $T$}
    \STATE{Sample $|B|$ instances from $\hat{X}$ and fetch the corresponding records in $R$;}
    \STATE{Update $\Theta$ according to (\ref{SGD2})}

\ENDFOR
    \end{algorithmic}
\end{algorithm}

\subsection{Sparse Regression Tree-based Valuation}
Decision tree is superiority in its interpretability among numerous machine learning models. It has been investigated in deep learning scenarios. Frosst and Hinton~\cite{DAN2017} leveraged the neural network to learn decision trees that mimic the underlying function discovered by the neural network.
Wan et al.~\cite{NBDT} replaced the decision layer of a deep neural network with a differentiable decision tree and a surrogate loss. However, the models constructed by these two methods are only transparent. In addition, they are not suitable for tabular data\footnote{The input of the MLP is tabular data.}. Zhou et al.~\cite{ITS2023} proposed neural regression tree~(NRT), which is based on a neural classification model Deep Neural Decision Tree (DNDT)~\cite{DNDT}, is proposed to investigate the weighting mechanisam in deep learning. Compared with the fixed structure of DNDT, the structure of NRT can be optimized with gradient descent in training, thereby yielding a more effective and interpretable model. The sparse regression tree~(SRT) is built based on improvements on NRT.

\subsubsection{Sparse Regression Tree}
There are three limitations for NRT as follows:
\begin{itemize}
    \item The space complexity is still high. Let $K$ and $L$ be the number of feature dimension and the average split of feature dimension in NRT. The space complexity is approximately $O(K^L)$, which is quite large for even small values of $K$ and $L$~(e.g., 10).  
    \item The growth and pruning strategy is complex. In NRT, two hyper-parameters are introduced to judge whether a new cut point is required for growth and whether two cut points should be merged for pruning.
    \item The interpretability in terms of features is weak. The cut points for features in NRT are generated or merged according to the optimizing rules. Undoubtedly, some cut points have no exact means. Consequently, although many tree-based rules can be obtained from NRT, these cut points impair the interpretability of the generated rules.  
\end{itemize}

To deal with these three limitations, an improved neural regression tree, namely, sparse regression tree~(SRT), is proposed. The first stage initialized a SRT, which consists of three steps. Firstly, each feature dimension is discretized into up to $L$ folds by leveraging a clustering technique, e.g., Kmeans. Subsequently, each feature value is replaced by the center of the cluster it locates. {There are two merits for this step.} Consequently, as the clusters in one dimension, semantics can be assigned to each cluster. For example, if three clusters are generated for a feature, then ``small", ``medium", and ``large" can be assigned to the three clusters according to the ordering of their cluster centers. Secondly, the cut points can be selected only from the cluster boundaries of each feature. Resultantly, the computational complexity can be greatly reduced. Take the loss feature~(i.e., $u_{\cdot, 1}$) as an example. Assuming that the cluster centers obtained by Kmeans are $\{\Tilde{c}^{1}_1, \cdots, \Tilde{c}^{L}_1\}$, for an instance $u_{m}$, its loss feature $u_{m,1}$ will be transformed to value as follows:
\begin{equation} 
    {\Tilde{u}_{m,1} = \Tilde{c}^{l^*}_1},\quad \text{s.t. } l^* = arg\min_{l \in [1,\cdots, L] } |u_{m,1}-\Tilde{c}^{l}_1|.
\end{equation}
We denote that $\pi[\Tilde{u}(m,i)]$ is a $L$-dimensional one-hot vector with the dimension of $l^*$ for $\Tilde{u}_{m,i}$ equaling one.

Secondly, the Kronecker product is calculated to determine all possible nodes of the tree as
follows:
\begin{equation}
\begin{aligned}
\boldsymbol{z}=\pi\left[{U}{(\cdot, 1)}\right] \otimes \pi\left[{U}{(\cdot, 2)}\right] \otimes \cdots \otimes \pi\left[{U}{(\cdot, 10)}\right].
\end{aligned}\label{kronecker}
\end{equation}
where   
$\boldsymbol{z}_{m} \in \mathrm{R}^{10L}$ in $\boldsymbol{z}$ is a one-hot vector, which is the index of the leaf node where $u_{m}$ arrives. According to Eq.~(\ref{kronecker}), the number of all possible leaf nodes is $L^{10}$, which grows very quickly when $L$ increases. However, the upper bound of the number of the leaf node which contains at least one sample is the number of instances\footnote{As the features are discretized according to the clustering results, the actual number of lead nodes is usually much smaller than that of the instance number.}. Therefore, when the number of instances is much smaller than $L^{10}$, sparse Kronecker product~\cite{SKC2021} can be used.

Thirdly, each leaf node is assigned with a value which is exactly the data value~(i.e., $\beta_i$). The whole structure of the initial constructed sparse regression tree is shown in Fig.~3. The weights $\beta$ are the regression coeffeicients and also the data values. The next stage is to optimize the tree in terms of both the structure and the weights $\beta$. 

To avoid the complex growth and pruning scheme employed in NRT, this study only adopts the growth scheme. Specifically, the tree is initialized based on a small number of bins $L'$~($L' < L$) for each feature dimension. For example, if $L'$ is set as three, then each feature dimension is divided into three regions. Accordingly, the space for $z_m$ in Eq.~(\ref{kronecker}) is $R^{3L}$ rather than $R^{10L}$. For each dimension, $L'$ regions are selected with equal intervals. Subsequently, the tree grows by adding new cut points. We consider the adding of $C$ cut points in each step.

In conventional regression tree, a widely-used split criteria is the ground-truth target variance of samples in each current leaf node. There is no ground-truth targets in our study. The $C$ cut points are selected as follows. Each selected $\beta_n$ corresponds to a leaf node. We can select top-$C$ $\beta$s according to the variances of the weighted dispersion of their corresponding leaf nodes:
\begin{equation} 
     \mathcal{D}(\beta_n) = \sum_{j:z_{j,n} = 1} |\beta_n|\times||u_j - c_n||_2^2,\label{pbeta}
\end{equation}
where $c_n$ is the cluster center of the $n$th leaf node. $|\beta_n|$ is used as the weight due to the consideration that the lead nodes with large $\beta$ values are prioritized for split.

The larger the value of $pg(\beta_n)$ is, the larger the error contributed by the corresponding subset. Subsequently, each of the $C$ selected leaf nodes is split into two new leaf nodes if at least one feature dimension contains at least one cut-off point that is not used. Then all the unused cut-off points within in region corresponding to the leaf node are taken turns to be tested. The unused cut-off point that can yield the new leaf nodes with the largest variance reduction in the two new leaf nodes is selected and added into the tree.

\subsubsection{Interpretable Valuation}
When SRT is used for data valuation, (\ref{LV-MLP}) becomes
\begin{equation} 
        arg\underset{\Theta}{\min}||R- \hat{X}f_{SRT}(U;\Theta)||_2^2,\label{LV-SRT}
\end{equation}
where $f_{SRT}$ represents the SRT model. Although MSE-based regression has an analytical solution, the solving relies on Matrix inversion and thus the computational complexity is unacceptable when the size is large. Therefore, this study still adopts the gradient descent~(GD)-based solving with the following updating formula: 
       \begin{equation} 
        \begin{aligned}
            \beta^{t+1} &= \beta^{t} - \eta\frac{\partial ||R- \hat{X}f_{SRT}(U;\beta)||_2^2}{\partial \beta}|_{\beta^{t}}\\
            &=\beta^{t}-2\eta[\hat{X}^T\hat{X}\beta^t - \hat{X}^TR] 
        \end{aligned},\label{SGD1}
       \end{equation}
where $\eta$ is the learning rate. $\beta^0$ can be set as the solution from the optimal $\beta$ solved before the current tree growth. The whole algorithm is called SRTbV and the algorithmic steps are presented in Algorithm 2.
\begin{algorithm}[t]
\small
    \caption{SRTbV}
    \label{alg1}
    \textbf{Input}: Original corpus $D= \{X, Y\}$, sampling rate $p$, $M$, $L$, $L'$, $K$, batch size $|B|$, \#epochs $T$, $\eta$.\\
       \textbf{Output}: Trained valuation tree SRT($\cdot$).
    \begin{algorithmic}[1]\vspace{-0.0in}
\STATE {Initialize $\Theta^{0}$;}
\FOR{$m = 1$ to $M$}
    \STATE{Sample a subset $D_m$ from $D$ with probability $p$;}
    \STATE{Train a model based on $D_m$;}
    \STATE{Evaluate the performance of the model and record it as $R_m$;}
\ENDFOR
\STATE {Calculate new feature matrix $\hat{X}$;}
\STATE {Extract the value features $U$ based on an independent learning process on raw samples and labels;}
\STATE {Cluster each dimension of $U$ into up to $L$ clusters via Kmeans;}
\STATE {Discretize each dimension of each instance in $U$ with the closest cluster center. The discretized set is termed as $\hat{U}$;}
\STATE {Split each dimension into $L'$ regions;}
\STATE {Calculate and storage the sparse Kronecker product for
 \\the possible leaf nodes;}
\STATE {Randomly divide $\{\hat{X}, R, U\}$ into five folds;}
\FOR {$t = 1$ to $T$}
    \STATE{Sample $|B|$ instances from $\hat{X}$ and fetch the corresponding records in $R$;}
    \STATE{Update $\Theta$ according to (\ref{SGD2});}

\ENDFOR
    \end{algorithmic}
\end{algorithm}

Once the SRT is learned, interpretable rules can be derived based on the rule-extraction approach utilized by Zhou et al.~\cite{ITS2023}. Note that 
SRT is a flat tree with no hierarchical structures among middle nodes as conventional decision trees. To visualize the regression
tree, information gain used in conventional decision trees is leveraged to construct the hierarchical order of the valuation characteristics. The valuation characteristic with the highest information gain
is placed in the first layer. The rest layers are determined according to the ordering of the characteristics' information gain. Specifically, the achieved values~(i.e., $\beta_n$s) are clustered into five groups whose values are from low to large using k-means to calculate the information gain. These five groups are named ``quite low", ``low", ``moderate", ``large", and ``quite large". Then, we can construct a (hierarchical) tree for each value group. As a consequence, which and how characteristics determine the values of samples can be obtained.

%

\section{Experiments}
This section answers the question for the learnable and interpretable for data valuation. Firstly, the performances of our proposed valuation methods MLPbV and SRTbV are evaluated. If the performance is superior or comparable to existing classical methods, then we can conclude that a data valuation model can be learned. Secondly, if the rules learned by SRTbV can be exposed and understood, then we can conclude that data valuation is interpretable.

\subsection{Experimental Setup}
Our methodology is evaluated along three main axes. First, we evaluate our learning-based SV estimator and compare it to classical methods especially the AME method. As previously stated, there are different paradigms for data valuation, this study follows the SV-based measurement paradigms. Nevertheless, as discussed in Section 4, our methodology can be extended to some other paradigms. Second, two downstream tasks of data valuation, namely, point removal and addition, are leveraged to evaluate our methodology. Most previous studies choose this task for performance evaluation\footnote{Two other popular downstream tasks, namely, noisy label detection and noisy feature detection, are not considered in this study. The reason lies in that the most well-known platform OpenValue chooses to apply clustering to the data valuation scores to identify noisy samples. In our point of view, this technique is inappropriate for methods such AME.}. Thirdly, the valuation rules in the learned sparse regress tree are extracted and presented. If the valuation rules are in accordance with existing domain knowledge for data valuation in machine learning, then the interpretability of SRTbV is verified.

The following classical marginal contribution-based methods are compared in the experiments along the first and the second axes:
\begin{itemize}
    \item AME~\cite{MIE2022}: The AME gives the exact value of the expected (average) marginal effect of adding a data point to a subset of the training data. The AME-LASSO is a practical version of AME, which estimate the exact value of AME by reducing the number of subsets and using a Lasso regression instead of the MSE. The AME-LASSO is compared with our method. 
    \item DataShapley~\cite{DSE2019}: The DataShapley estimates the Shapley value of data using Monte Carlo and a gradient-based implementation of DataShapley is developed for large-scale benchmarks.
    \item KNNShapley~\cite{ETD2019}: The KNNShapley approximates the exact Shapley value based on K-nearest neighbors' data shapley scores, i.e., the KNN utility. It measures the boost of the likelihood that KNN assigns the correct label to each test data point. 
    \item Volume-based Shapley~\cite{VFA2021}: The Volume-based Shapley decouples the dependence of the shapley value on the validation set by using a input-concerned volume-based valuation metric instead of the typical utility score evaluating on the validation set. An implementation of the volume-based Shapley, namely VSV, is compared with our method. 
    \item BetaShapley~\cite{BSA2022}: The BetaShapley estimates the exact shapley value by using a weighted mean of the marginal contributions instead of the averaged marginal effect. The employed weights are calculated using the Beta function. 
    \item DataBanzhaf~\cite{DBA2023}: The DataBanzhaf introduces the Banzhaf values as the weights on the marginal contributions. The Banzhaf values are robust to noisy model performance scores.
    \item LAVA~\cite{LDV2023}: The LAVA values individual data regardless of the design choices of the underlying learning algorithm by introducing a class-wise Wasserstein distance.
\end{itemize}
A newly proposed method, namely, Data-OOB~\cite{DOE2023}, is also compared as its superiority performance. Data-OOB uses the Out-of-bag~(OOB) estimate to measure the quality of data. Data-OOB uses the averaged goodness of several weak learners instead of the averaged marginal effect. 

\begin{figure}[t] 
    \centering \includegraphics[width=1\linewidth]{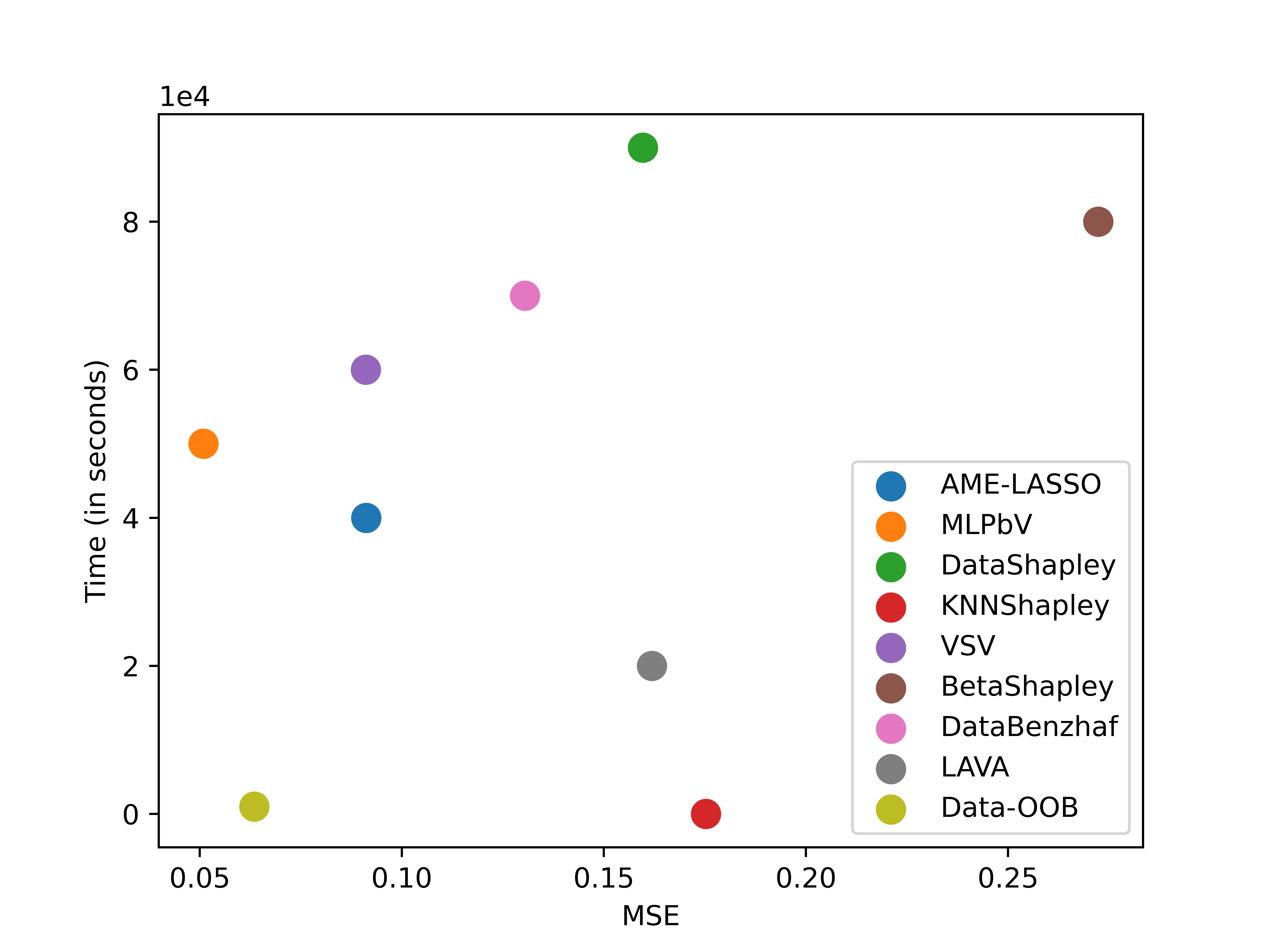}     \vspace{-0.3in}
    \caption{The comparison results on Shapley value estimation on CIFAR10.}
    \label{fig1}
     \vspace{-0.02in}
\end{figure}

\begin{figure}[t] 
    \centering \includegraphics[width=1\linewidth]{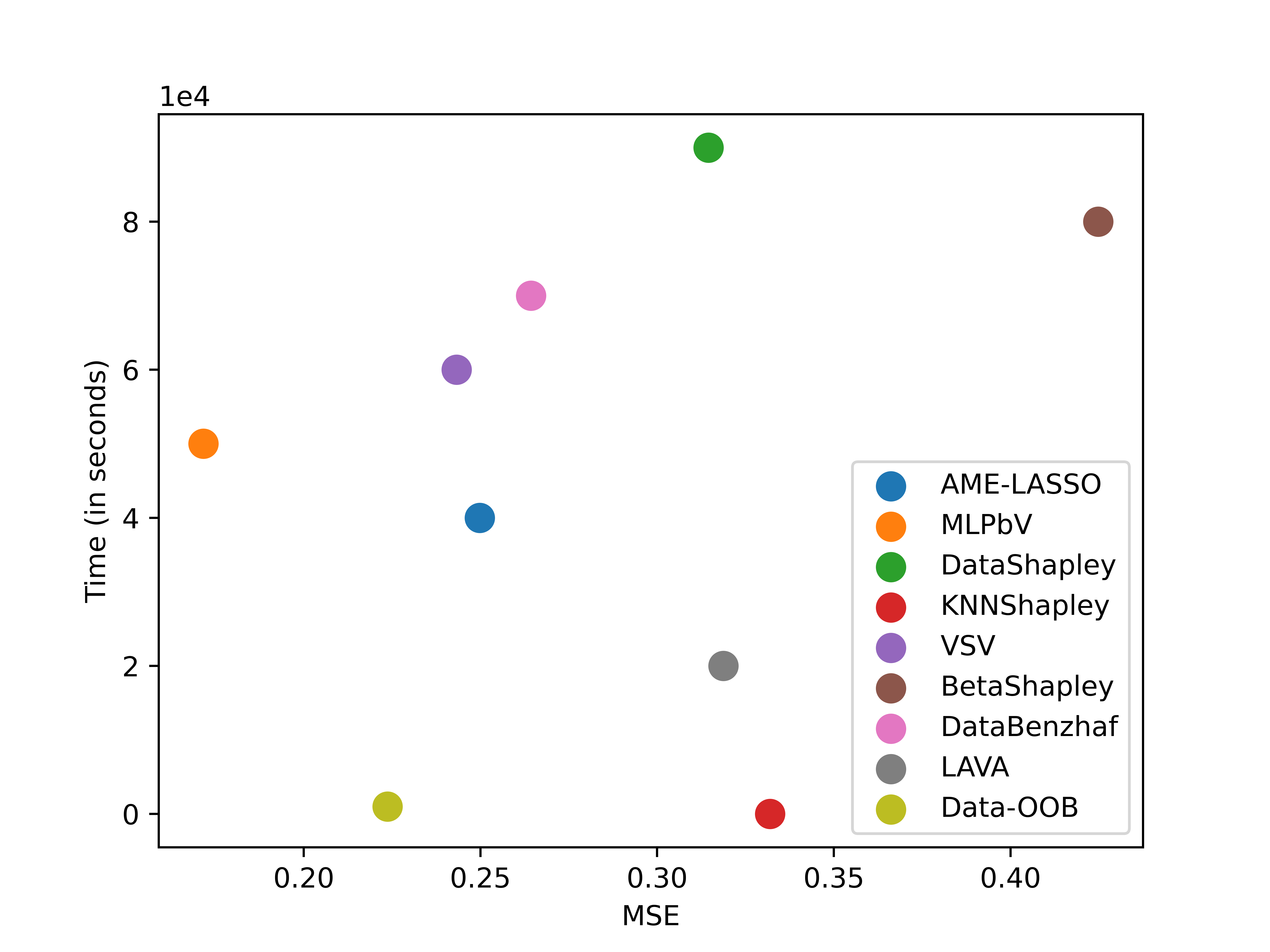}     \vspace{-0.3in}
    \caption{The comparison results on Shapley value estimation on CIFAR100.}
    \label{fig1}
     \vspace{-0.02in}
\end{figure}

\begin{figure}[t] 
    \centering \includegraphics[width=1\linewidth]{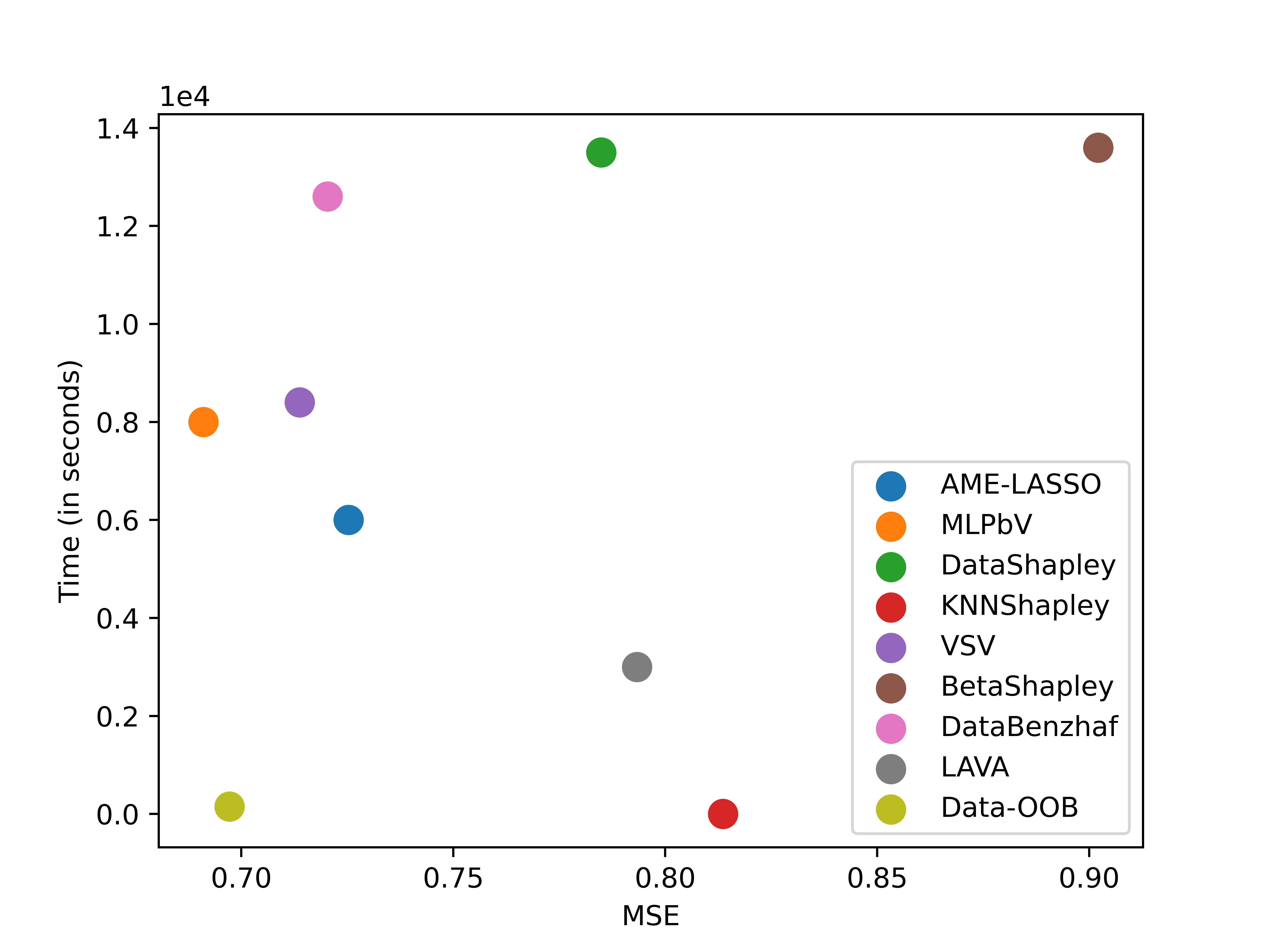}     \vspace{-0.3in}
    \caption{The comparison results on Shapley value estimation on BBC.}
    \label{fig1}
     \vspace{-0.02in}
\end{figure}

\begin{figure}[t] 
    \centering \includegraphics[width=1\linewidth]{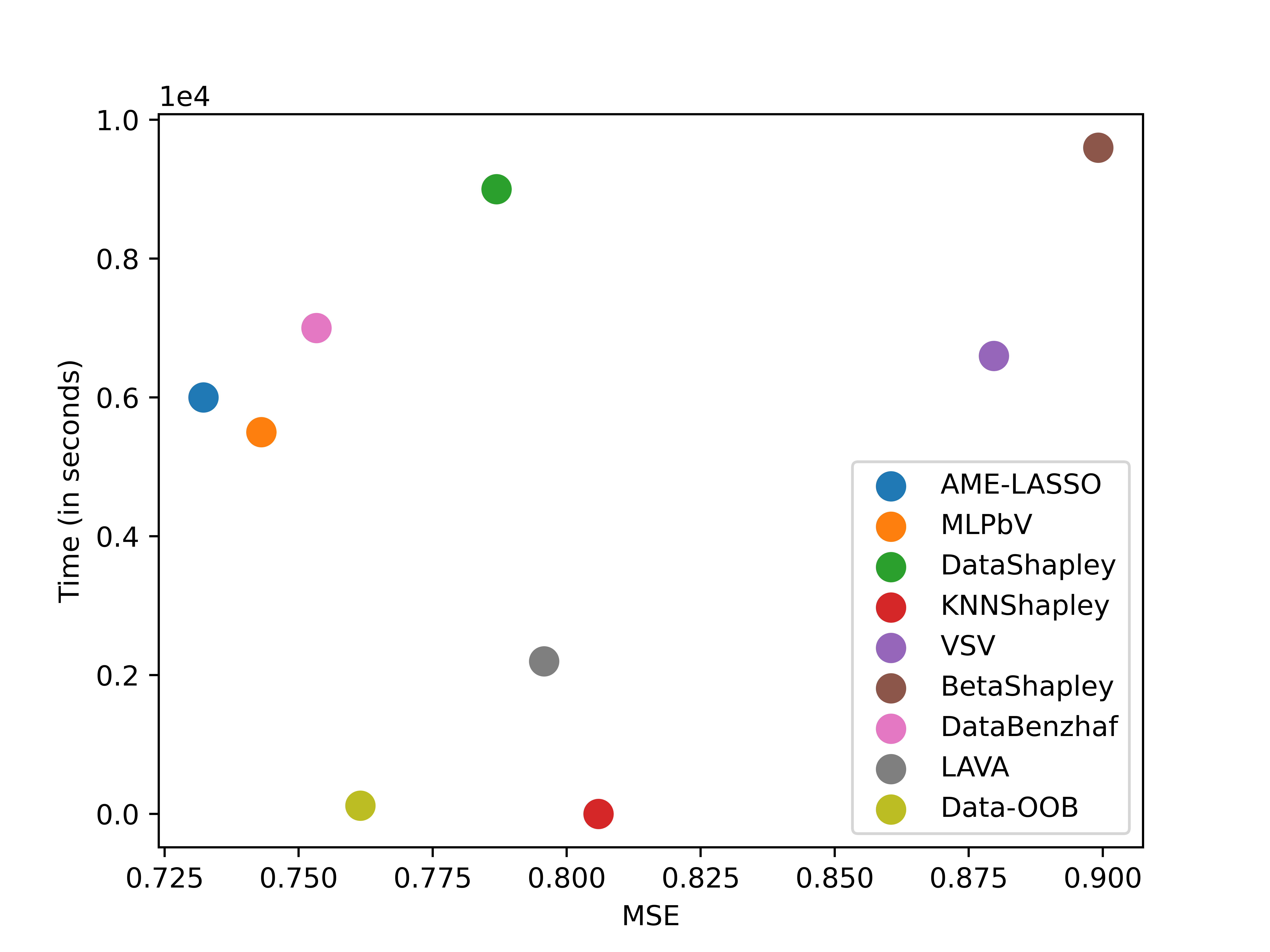}     \vspace{-0.3in}
    \caption{The comparison results on Shapley value estimation on IMDB.}
    \label{fig1}
     \vspace{-0.02in}
\end{figure}

As this study takes deep learning as the application scenario, five benchmark data sets which are usually used in DNN-based classification are leveraged in our experiments:
\begin{itemize}
    \item CIFAR10~\cite{Krizhevsky32}: CIFAR10 is an image data set. It is composed of 50,000 training data and 10,000 test data belonging to 10 classes. 
    \item CIFAR100~\cite{Krizhevsky32}: CIFAR100 is an image data set. It is composed of 50,000 training data and 10,000 test data belonging to 100 classes. 
    \item IMDB~\cite{LWC2011} IMDB is a text data set. It consists of 50,000 training data and 500 test data belonging to 2 classes. 
    \item BBC~\cite{PST2006}: BBC is a text data set. It consists of 2225 training data and 500 test data belonging to 5 classes.
    \item ImageNet~\cite{ImageNet0FeifeiLi}: ImageNet is a large-scale image data set including 1.28 million training data, 50,000 validation data, and 100,000 test data belonging to 1,000 classes.
\end{itemize}

The evaluation metrics are detailed in the following subsections. On the two image datasets, ResNet50 is used as the backbone networks. On the two text datasets, DistilBERT is used as the backbone networks. The warm-up setting described in~\cite{MIE2022} is used for both . The detailed settings for the hyper-parameters are described in the corresponding experiments.

\begin{figure*}[t] 
    \centering \includegraphics[width=1\linewidth]{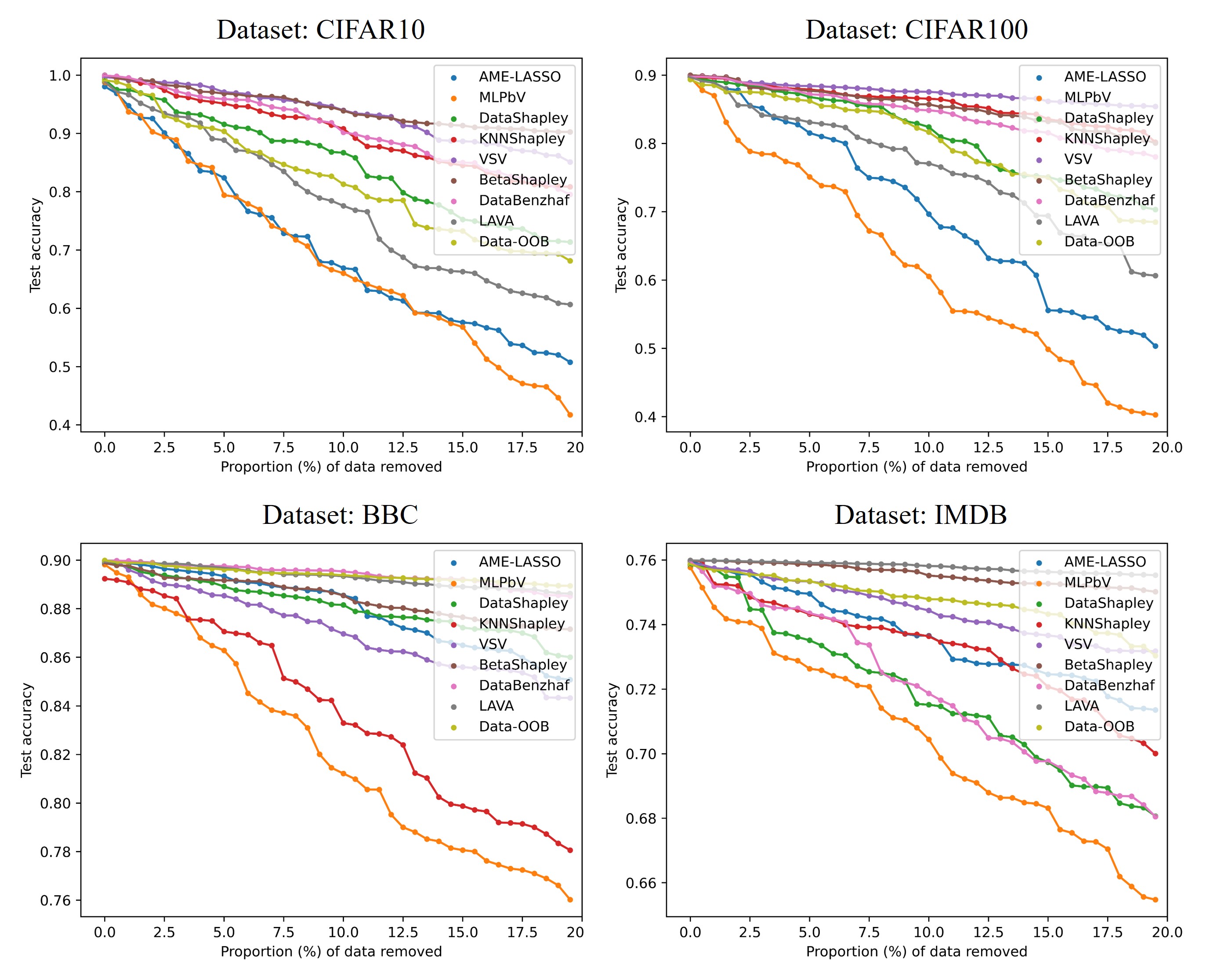}     
    \caption{\textbf{Point removal experiment.}  Test accuracy curves as a function of the most valuable data points added. Lower curve indicates better data valuation algorithm.--cifar10}
    \label{fig1}
\end{figure*}

\subsection{Results on Shapley Value Estimation}
To obtain the ground-truth Sharpley value, we refer to the theoretical basis of AME, which claims that when the number of sampled subsets $M$ approaches to $+\infty$, the value by solving the regression problem approaches to the ground-truth Sharpley value. In this study, $M$ is set as the size of the entire training corpus.

Two evaluation metrics are leveraged in this part. The first is the MSE score between the predicted values and the ground-truth values. The lower the MSE score, the better the achieved performance. The second is the normal Kendall-tau distance. 

For the two backbone networks on CIFAR10 and CIFAR100, the hyper-parameters are set as follows.

\begin{figure}[t] 
    \centering \includegraphics[width=1\linewidth]{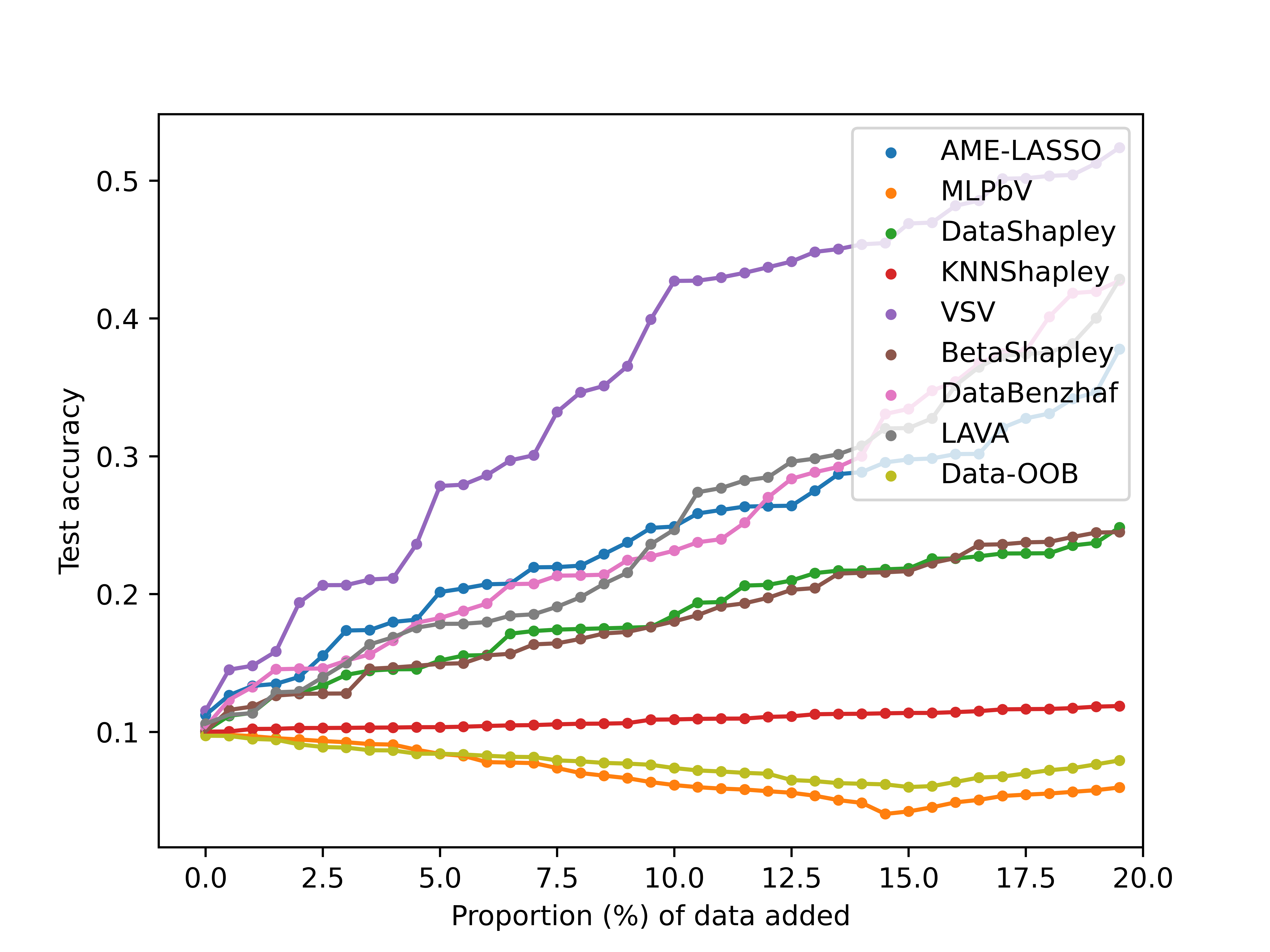}     \vspace{-0.3in}
    \caption{\textbf{Point addition experiment.}  Test accuracy curves as a function of the least valuable data points added. Lower curve indicates better data valuation algorithm. --cifar10 }
    \label{fig1}
     \vspace{-0.02in}
\end{figure}

\begin{figure}[t] 
    \centering \includegraphics[width=1\linewidth]{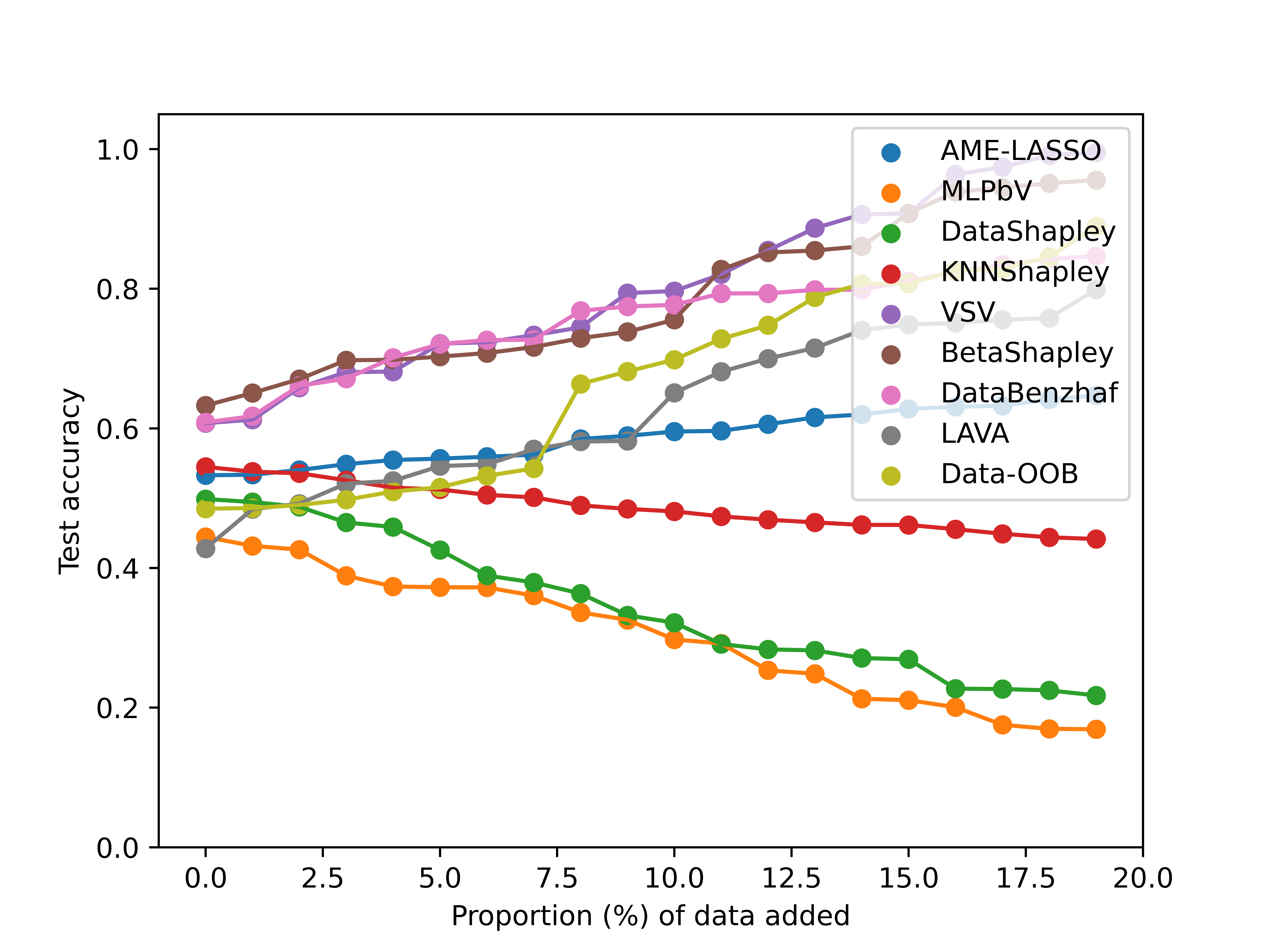}     \vspace{-0.3in}
    \caption{\textbf{Point addition experiment.}  Test accuracy curves as a function of the least valuable data points added. Lower curve indicates better data valuation algorithm.--cifar100 }
    \label{fig1}
     \vspace{-0.02in}
\end{figure}

\begin{figure}[t] 
    \centering \includegraphics[width=1\linewidth]{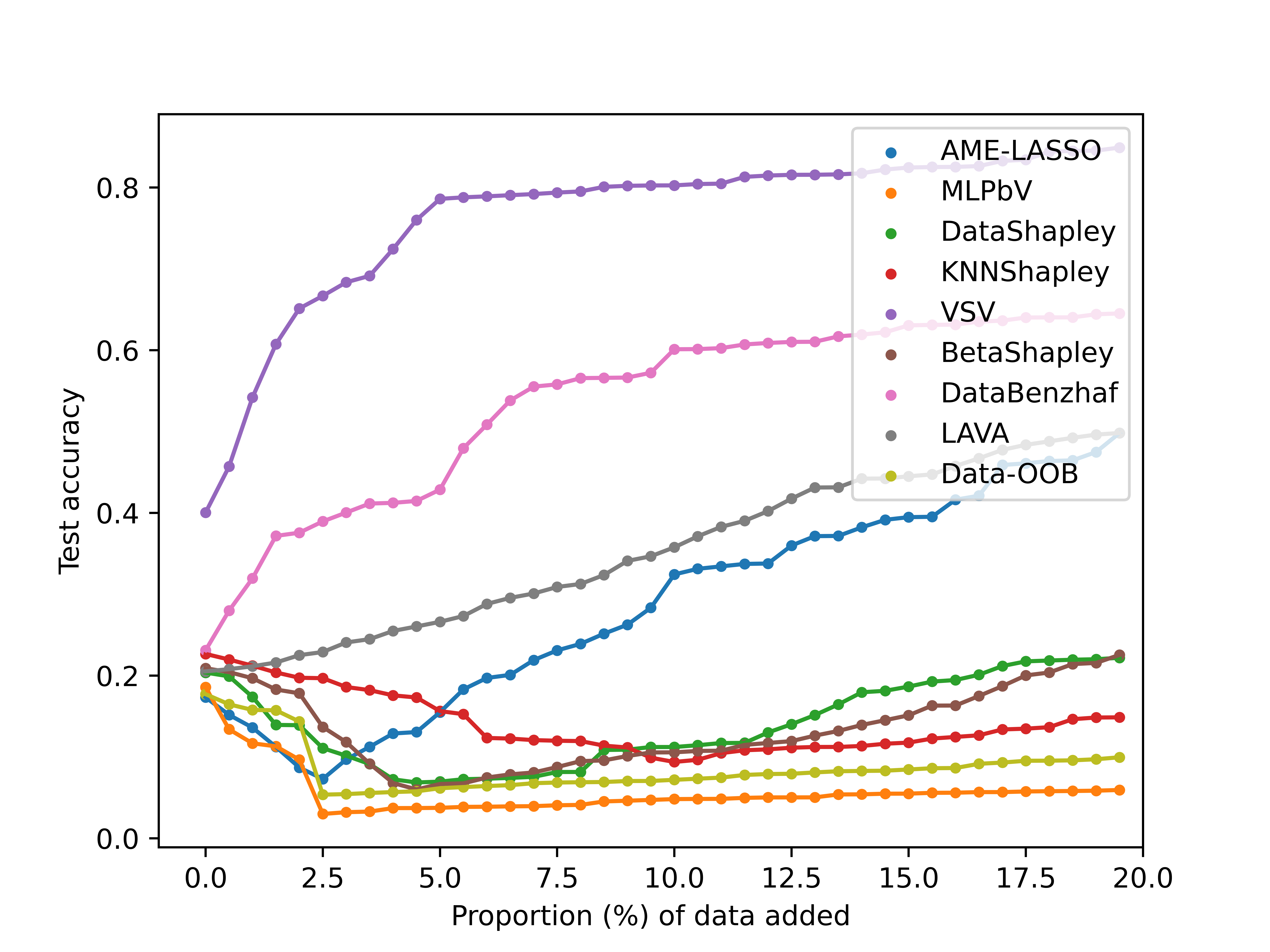}     \vspace{-0.3in}
    \caption{\textbf{Point addition experiment.}  Test accuracy curves as a function of the least valuable data points added. Lower curve indicates better data valuation algorithm. --bbc }
    \label{fig1}
     \vspace{-0.02in}
\end{figure}

\begin{figure}[t] 
    \centering \includegraphics[width=1\linewidth]{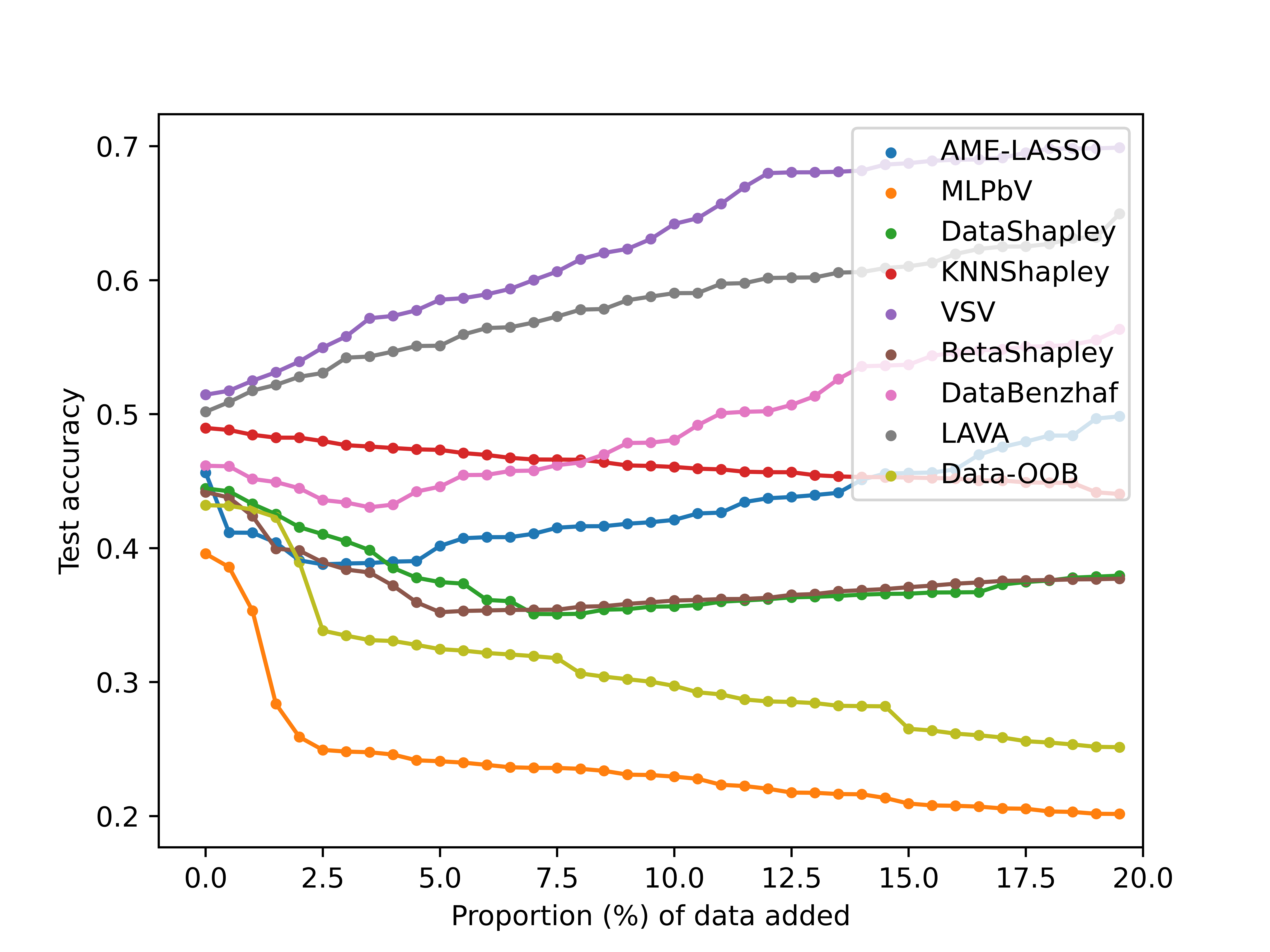}     \vspace{-0.3in}
    \caption{\textbf{Point addition experiment.}  Test accuracy curves as a function of the least valuable data points added. Lower curve indicates better data valuation algorithm. --imdb }
    \label{fig1}
     \vspace{-0.02in}
\end{figure}

\subsection{Results on Point Removal and Addition}
Two downstream tasks are implemented in this section. We perform the point removal experiment and the point additional experiment following the experimental settings in~[]. For the point removal experiment, training data are ranked according to their values in a descending order. Data with high values are removed one by one from the training set, and at most $K=\lfloor  0.2\times N\rfloor$ data are removed from the training set. For the point additional experiment, training data are ranked according to their values in an ascending order. The training data with lowest values are added in an empty set in order, and the set is used as the training set. At most $K=\lfloor  0.2\times N\rfloor$ data are added in the empty set. For both experiments, we evaluate the model trained with the modified training set on the test set and observe the variation of the test accuracy.




\section{Conclusion}
This study has investigated the learnable and explainable issues for data valuation within deep learning tasks. First, an MLP-based valuation framework is presented. This framework can learn an MLP-based valuation model on the basis of a number of features which can reflect the values of each training sample. Second, the MLP model is replaced by a differential regression tree which can be trained with similar steps used for existing neural regression tree. As a consequence, valuation rules can be derived from the trained differential regression tree. Extensive experiments and benchmark data valuation datasets verify the effectiveness of the proposed methodology.

\bibliographystyle{IEEEtran}
\bibliography{ref}

%

\end{document}